\documentclass[runningheads]{llncs}
\usepackage[T1]{fontenc}
\usepackage{graphicx}
\usepackage{booktabs}
\usepackage{multirow}
\usepackage{float}
\usepackage{amsmath,amssymb}
\usepackage{caption}
\usepackage[hidelinks]{hyperref}
\usepackage{adjustbox}
\usepackage{subcaption}
\usepackage[table]{xcolor}
\begin{document}
\title{MedDiME: Efficient Latent Diffusion with Adaptive Masking for Medical Counterfactual Generation}
\titlerunning{MedDiME}
%
\author{Yan Zeng\thanks{Equal contribution.} \and
Changlu Guo\textsuperscript{\thefootnote} \and
Anders Nymark Christensen \and
Morten~Rieger~Hannemose \and
Anders Bjorholm Dahl}
\authorrunning{Y. Zeng et al.}
\institute{Department of Applied Mathematics and Computer Science,\\
Technical University of Denmark\\
\email{\{s242652,chagu,anym,mohan,abda\}@dtu.dk}}
\maketitle              
%

\begin{abstract}
Medical counterfactual generation modifies images to change model predictions for interpretability. However, existing diffusion-based approaches are often prohibitively slow and memory-intensive, making them difficult to apply in high resolution settings. Moreover, existing masking strategies are tightly coupled with pixel-space representations, making them incompatible with latent-space diffusion editing. To address these challenges, we propose MedDiME, a latent-space classifier-guided diffusion framework that reduces computational and memory overhead while introducing a latent-compatible, gradient-driven adaptive masking mechanism for spatially precise medical counterfactual generation. Extensive experiments demonstrate that MedDiME achieves high-quality counterfactual generation with significant efficiency gains compared to prior classifier-guided diffusion baselines, achieving up to \textbf{40$\times$} faster inference and \textbf{13$\times$} lower peak GPU memory usage.


\end{abstract}

\section{Introduction}
Deep learning models in medical imaging often operate as black boxes, limiting clinical trust and practical deployment~\cite{haselhoff2024gaussian}. Counterfactual generation modifies regions relevant to model decisions to alter predictions~\cite{augustin2022diffusion,singla2019explanation}. Fundamentally, it aims to achieve decision-driven semantic reversal rather than pixel-level exact reconstruction. In medical imaging, such editing requires precise control over spatially localized, decision-relevant regions while preserving overall structural plausibility. However, early generative counterfactual methods in medical imaging suffer from limited controllability and instability, often producing semantically distorted or structurally inconsistent results~\cite{khorram2022cycle,pombo2023equitable,pahde2023reveal}.

\begin{figure}[t!]
  \centering
  \includegraphics[width=\textwidth]{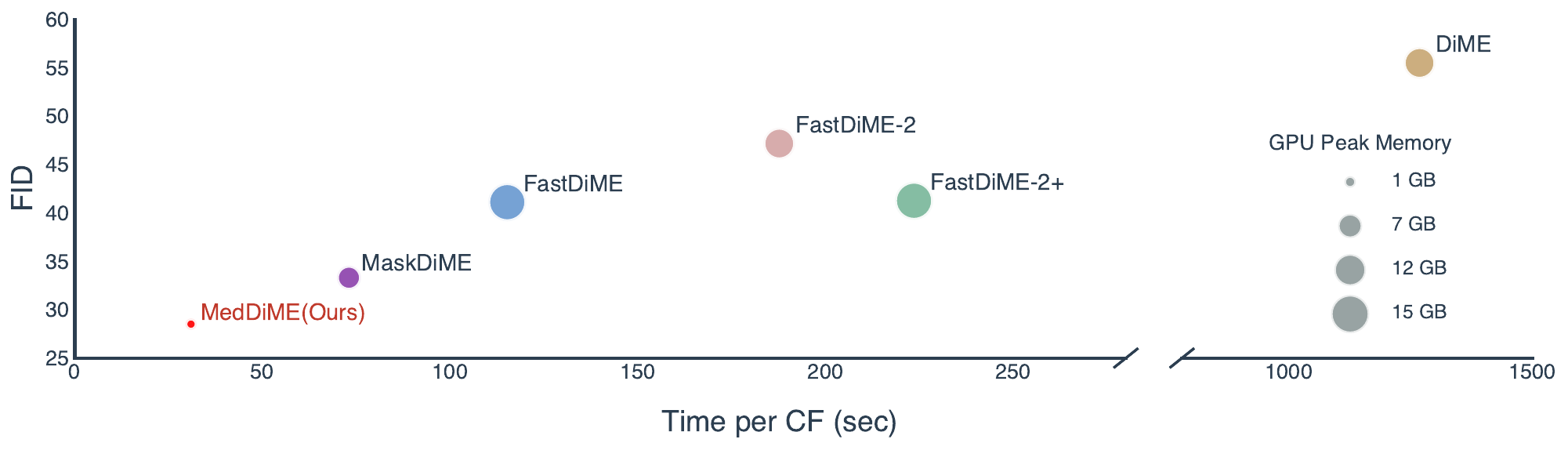}
 \caption{Quality--efficiency trade-off for counterfactual generation on ISIC2018 (Ruler--Non-Ruler).
FID is plotted against time per counterfactual (sec), with bubble size denoting peak GPU memory.
MedDiME dominates the lower-left region, achieving the best quality--efficiency--memory balance among all methods.}

  \label{fig:pipeline_efficiency}
\end{figure}
Recently, Denoising Diffusion Probabilistic Models~\cite{ho2020denoising,dhariwal2021diffusion} have emerged as a more stable alternative for counterfactual generation. Among them, DiME~\cite{jeanneret2022diffusion} incorporates classifier guidance during sampling but requires full denoising rollouts with global pixel-space guidance, resulting in high computational cost and spatially diffuse edits. To mitigate these issues, FastDiME~\cite{weng2024fast} improves efficiency by approximating the clean image with a single-step denoising operation and computing classifier guidance accordingly, while incorporating a self-optimized masking scheme to enhance edit locality. Its masking strategy is based on pixel-level differences rather than semantic or decision-level cues, which may misalign with decision-relevant regions and result in imprecise localization and over-editing~\cite{vath2023diffusion,melistas2024benchmarking}.
In contrast, MaskDiME~\cite{guo2026maskdime} constructs adaptive masks directly from classifier gradients, enabling the model to dynamically focus on decision-relevant regions and achieve more precise, spatially localized counterfactual edits. 

Although the aforementioned methods have achieved improvements in efficiency and spatial precision, medical image analysis in clinical practice typically relies on high-resolution inputs. However, existing counterfactual approaches based on pixel-space diffusion are predominantly validated on natural images or relatively low-resolution settings (e.g., $128\times128$ or $256\times256$ pixels). When extended to higher resolutions (e.g., $512\times512$ pixels and above), pixel-space diffusion incurs substantially increased computational and memory costs, as these scale with the number of pixels. This rapid growth in modeling complexity poses significant challenges for high-resolution medical counterfactual generation.

Latent representations compress pixel redundancy, directing diffusion toward decision-relevant structural changes consistent with the semantic reversal objective in counterfactual generation. Operating in this lower-dimensional space also reduces the computational and memory burden introduced by high-resolution inputs~\cite{rombach2022high,luu2025visual,jeanneret2024text}. However, migrating classifier-guided diffusion editing to the latent space is not a straightforward substitution~\cite{yeganeh2025latent,kazimi2025explaining}. Unlike pixel space, latent representations lack explicit spatial correspondence, rendering masking strategies based on pixel differences or image-space cues conceptually incompatible. 

In response to these limitations, we propose MedDiME, a latent-space framework for classifier-guided medical counterfactual generation that integrates adaptive gradient-driven spatial masking. Our contributions are threefold:
(1) We conduct classifier-guided diffusion for counterfactual generation entirely in latent space, demonstrating that high-resolution medical image editing can be effectively achieved within a lower-dimensional representation while maintaining high counterfactual quality and significantly improving computational efficiency, as shown in Fig.\ref{fig:pipeline_efficiency}. (2) Rather than directly transferring pixel-space masking, we develop a latent-compatible classifier-gradient-driven adaptive masking mechanism that redefines spatial guidance within the latent representation for precise, spatially localized editing. (3) We show that MedDiME not only significantly improves computational efficiency, but also achieves state-of-the-art or competitive performance across multiple medical image counterfactual explanation tasks.

\section{Methodology}
\subsection{Framework overview}
\begin{figure}[t]
  \centering
  \includegraphics[width=\textwidth]{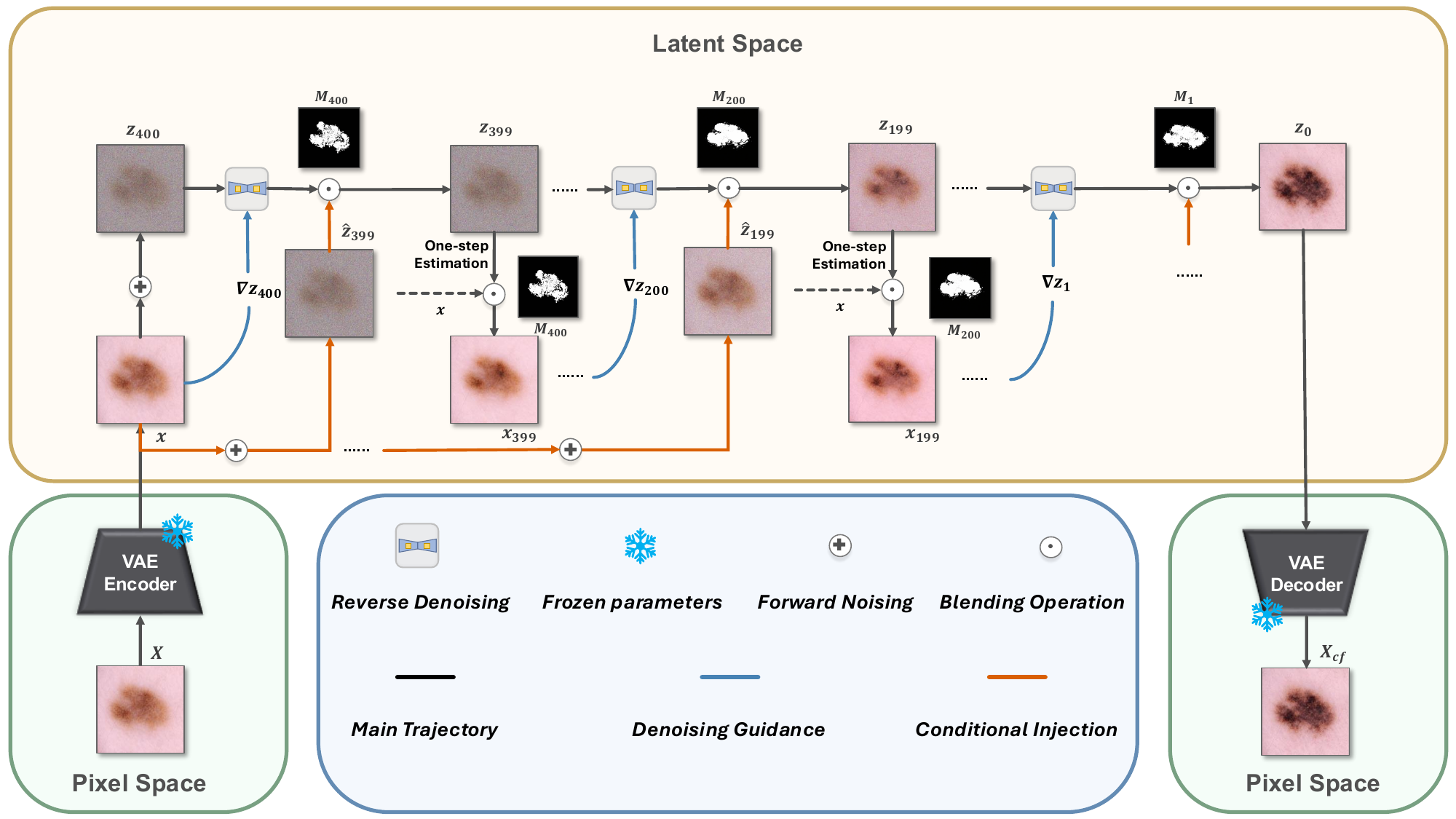}
\caption{\textbf{Overview of the MedDiME framework.}
Latent states are decoded only for visualization; all operations are conducted
in the latent space.}

  \label{fig:framework}
\end{figure}

Building upon the classifier-guided diffusion paradigm of DiME~\cite{jeanneret2022diffusion}, we introduce an entirely latent-space framework for medical counterfactual generation (Fig.~\ref{fig:framework}).

Given an input image $X$, we first encode it into a latent representation $x$
using a pretrained VAE encoder.
The latent representation is then diffused forward to a predefined timestep $\tau$
($1 \leq \tau \leq T$), yielding a noisy latent variable $\hat{z}_\tau$.
We initialize the reverse diffusion process with $z_\tau = \hat{z}_\tau$ and
iteratively refine the latent state under gradient guidance to obtain the final
counterfactual latent $z_0$.

At each reverse diffusion step $t$, we additionally maintain a latent estimate
$x_t$, which represents a denoised approximation of the current noisy latent
state and is used for gradient $\nabla_{z_t}$ computation (see Sec. \ref{sec:gradient} for details).
Specifically, the noisy latent variable is updated at each reverse diffusion
step as:
\begin{equation}
z_{t-1}
=
M_t \odot \mathcal{N}\!\big(
\mu_\theta(z_t) - \Sigma_\theta(z_t)\nabla_{z_t},
\ \Sigma_\theta(z_t)
\big)
+
(1 - M_t)\odot \hat{z}_{t-1},
\end{equation}
where $\odot$ denotes element-wise multiplication,
$\mu_\theta(\cdot)$ and $\Sigma_\theta(\cdot)$ are the predicted mean and
variance of the reverse diffusion process, and
$\hat{z}_{t-1}$ is obtained by applying the forward diffusion process to the
original latent representation.
This formulation restricts classifier-guided denoising updates to the regions
indicated by the adaptive mask $M_t$ (see Sec. \ref{sec:mask} for details), while preserving the original diffusion
trajectory outside the masked regions. After obtaining $z_{t-1}$, we apply a one-step denoising operation based on Tweedie's formula~\cite{efron2011tweedie}, following FastDiME~\cite{weng2024fast}, to estimate the corresponding clean latent representation.

\begin{equation}
\hat{x}_0^{(t-1)}
=
\frac{
z_{t-1} - \sqrt{1 - \bar{\alpha}_{t-1}}\,\epsilon_\theta(z_{t-1})
}{
\sqrt{\bar{\alpha}_{t-1}}
}.
\end{equation}
This estimate avoids full reverse denoising and reduces computation. The estimated latent state is then used to construct the next latent state:
\begin{equation}
x_{t-1}
=
M_t \odot \hat{x}_0^{(t-1)}
+
(1 - M_t)\odot x,
\end{equation}
where the adaptive mask $M_t$ restricts updates to the masked latent regions while preserving the original latent representation elsewhere. This masked latent state $x_{t-1}$ is subsequently used for gradient computation
and mask construction at the next reverse diffusion step.

The above procedure is repeated for all reverse diffusion timesteps until the
final latent state $z_0$ is obtained, which is then decoded by the VAE decoder to
produce the counterfactual image $X_{\mathrm{cf}}$.

\subsection{Gradient-Based Guidance in Latent Space}
\label{sec:gradient}
We follow DiME~\cite{jeanneret2022diffusion} and optimize a joint loss composed of a classification term $L_{\mathrm{class}}$, a perceptual term $L_{\mathrm{perc}}$, and an $L_1$ regularization term. The classification loss drives the latent toward the target class, while the perceptual and $L_1$ terms preserve structural consistency. The joint loss is formulated as:
\begin{equation}
L(x_t; y, x)
=
\lambda_c L_{\mathrm{class}}(C(y \mid x_t))
+
\lambda_p L_{\mathrm{perc}}(x_t, x)
+
\lambda_l L_{L_1}(x_t, x),
\tag{4}
\end{equation}
where $\lambda_c$, $\lambda_p$, and $\lambda_l$ control the relative importance of each term. 
We fix ${\lambda_c=5}$, $\lambda_p=50$, and $\lambda_l=1$ for all experiments. 
$C(y \mid x_t)$ denotes the predicted probability of the target class $y$ given the latent estimate $x_t$.

To enable gradient-based guidance during the reverse diffusion process, we
compute the gradient of the loss function $L$ with respect to the noisy
latent variable $z_t$.
To avoid backpropagation through the U-Net~\cite{ronneberger2015unet}, we follow  MaskDiME ~\cite{guo2026maskdime} and approximate the gradient with respect to $z_t$ as:
\begin{equation}
\nabla_{z_t}
=
\frac{s}{\sqrt{\bar{\alpha}_t}}
\nabla_{x_t} L(x_t; y, x),
\tag{5}
\end{equation}
where $s$ is a scalar factor controlling the overall guidance strength, and we set $s = 200$ by default in all experiments.

\subsection{Adaptive Masking in Latent Space}
\label{sec:mask}
To impose spatial constraints during latent-space diffusion, we train a classifier directly on latent representations and construct an adaptive mask at each reverse diffusion step based on its gradients. Unlike methods that define masks in the pixel space (e.g., FastDiME \cite{weng2024fast} and MaskDiME~\cite{guo2026maskdime}) or generate pixel-level masks and downsample them to the latent resolution (e.g., Blended Latent Diffusion \cite{luu2025visual}), our spatial constraint is constructed within the latent feature domain. This eliminates the need for pixel-to-latent mask projection and enables more accurate alignment with decision-sensitive regions in the latent representation.

Specifically, we first compute the spatial gradient induced by the
classification loss and approximately map it to the noisy latent variable
$z_t$:
\begin{equation}
\nabla^{\mathrm{class}}_{z_t}
=
\frac{1}{\sqrt{\bar{\alpha}_t}}
\nabla_{x_t} L_{\mathrm{class}}(C(y \mid x_t)).
\tag{6}
\end{equation}
Here, $x_t$ denotes the latent estimate at step $t$, and $C(\cdot)$ is a
latent-space classifier.

We then take the element-wise absolute value and average over the channel
dimension to obtain a scalar saliency map:
\begin{equation}
G_t
=
\left| \nabla^{\mathrm{class}}_{z_t} \right|_{\mathrm{avg}}
\in \mathbb{R}^{1 \times H \times W}.
\tag{7}
\end{equation}

The adaptive mask $M_t$ is defined by selecting the top-$k\%$ values of $G_t$: 
for each spatial location $(i,j)$, we set $M_t(i,j)=1$ if $G_t(i,j)$ is among the top-$k\%$, and 0 otherwise. 
We set $k=20\%$ for both ISIC2018 tasks and $k=30\%$ for CheXpert.
This mask confines classifier-guided updates to decision-relevant latent regions while preserving the original diffusion elsewhere. 
To enhance spatial coherence, we apply morphological dilation with a kernel size of 5.

\section{Experiments and Results}

\noindent \textbf{Datasets.} We evaluate MedDiME on three tasks, ranging from
editing salient but non-pathological visual structures to clinically meaningful
pathological semantic transformations.
CheXpert~\cite{irvin2019chexpert} and ISIC2018~\cite{codella2019skin}
provide chest X-ray and dermoscopic skin lesion images, respectively.
On both datasets, we consider counterfactual edits of non-pathological
display structures~\cite{pombo2023equitable,pahde2023reveal}, including
pacemakers in CheXpert and ruler markers in ISIC2018.
We further evaluate clinical counterfactuals
between Nevus and Melanoma on ISIC2018. All datasets are randomly split into training, validation, and test sets with a ratio of 7:2:1. The classifier and diffusion models are trained on the training set and validated on the validation set. Counterfactual generation and quantitative evaluation are conducted on the held-out test set.

\noindent \textbf{Evaluation Metrics.} As MedDiME follows the classifier-guided paradigm of DiME~\cite{jeanneret2022diffusion}, we restrict comparisons to methods within the same framework to ensure methodological consistency and fair evaluation. We follow the FastDiME~\cite{weng2024fast} and MaskDiME~\cite{guo2026maskdime} evaluation protocol and assess
counterfactual quality from three perspectives: realism,
proximity, and validity~\cite{vath2023diffusion,melistas2024benchmarking}.
Specifically, we report Fr\'echet Inception Distance (FID)~\cite{heusel2017gans}
for realism, pixel-level $L_1$ distance and SimSiam-based semantic similarity
(S$^3$)~\cite{chen2021exploring} for proximity, and Flip Rate (FR) together with
Mean Absolute Difference (MAD) for validity.

\noindent \textbf{Implementation Details.} MedDiME operates in latent space, whereas the baselines operate in pixel space~\cite{jeanneret2022diffusion,weng2024fast,guo2026maskdime}. 
For fair comparison, we standardize the core architectures and sampling configurations across methods, including a U-Net-based diffusion backbone~\cite{ronneberger2015unet}, a DenseNet121 classifier~\cite{huang2017densely}, the total number of diffusion steps ($T=1000$), and the starting noise level for counterfactual generation ($t=400$). For image-space methods, inputs are resized to a high resolution of $512\times512$ pixels. 
For latent-space modeling, we adopt the pretrained VAE from Stable Diffusion XL (SDXL)~\cite{rombach2022high} as a frozen encoder--decoder to map images into a $4\times64\times64$-dimensional latent representation and reconstruct the edited outputs. 
Unless otherwise specified, all remaining hyperparameters follow DiME~\cite{jeanneret2022diffusion}. 
All experiments are conducted on a single NVIDIA A100 80GB GPU.

\begin{figure}[t]
    \centering
    \begin{subfigure}[t]{0.5\textwidth}
        \includegraphics[width=\linewidth]{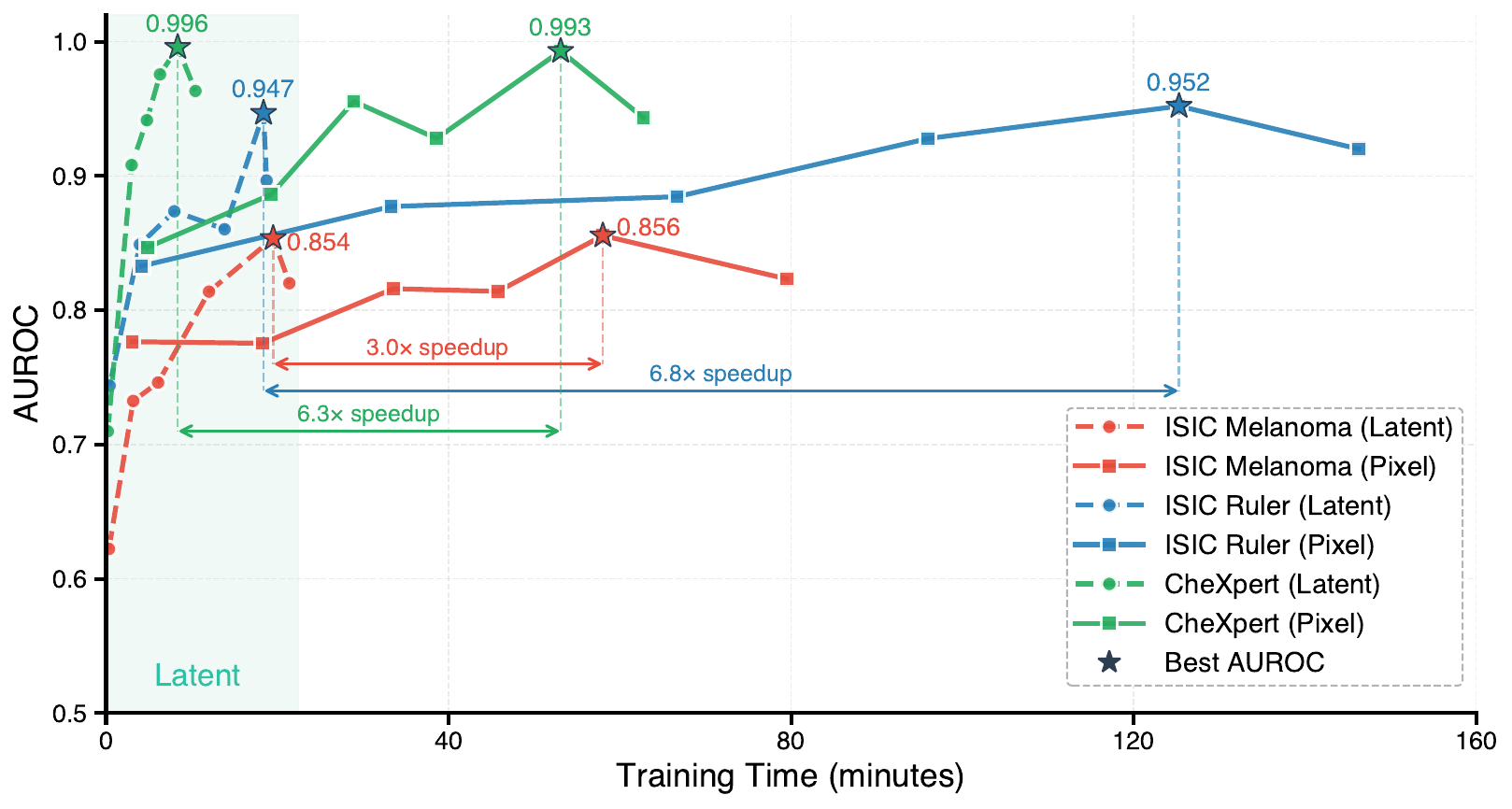}
    \end{subfigure}%
    \begin{subfigure}[t]{0.5\textwidth}
        \includegraphics[width=\linewidth]{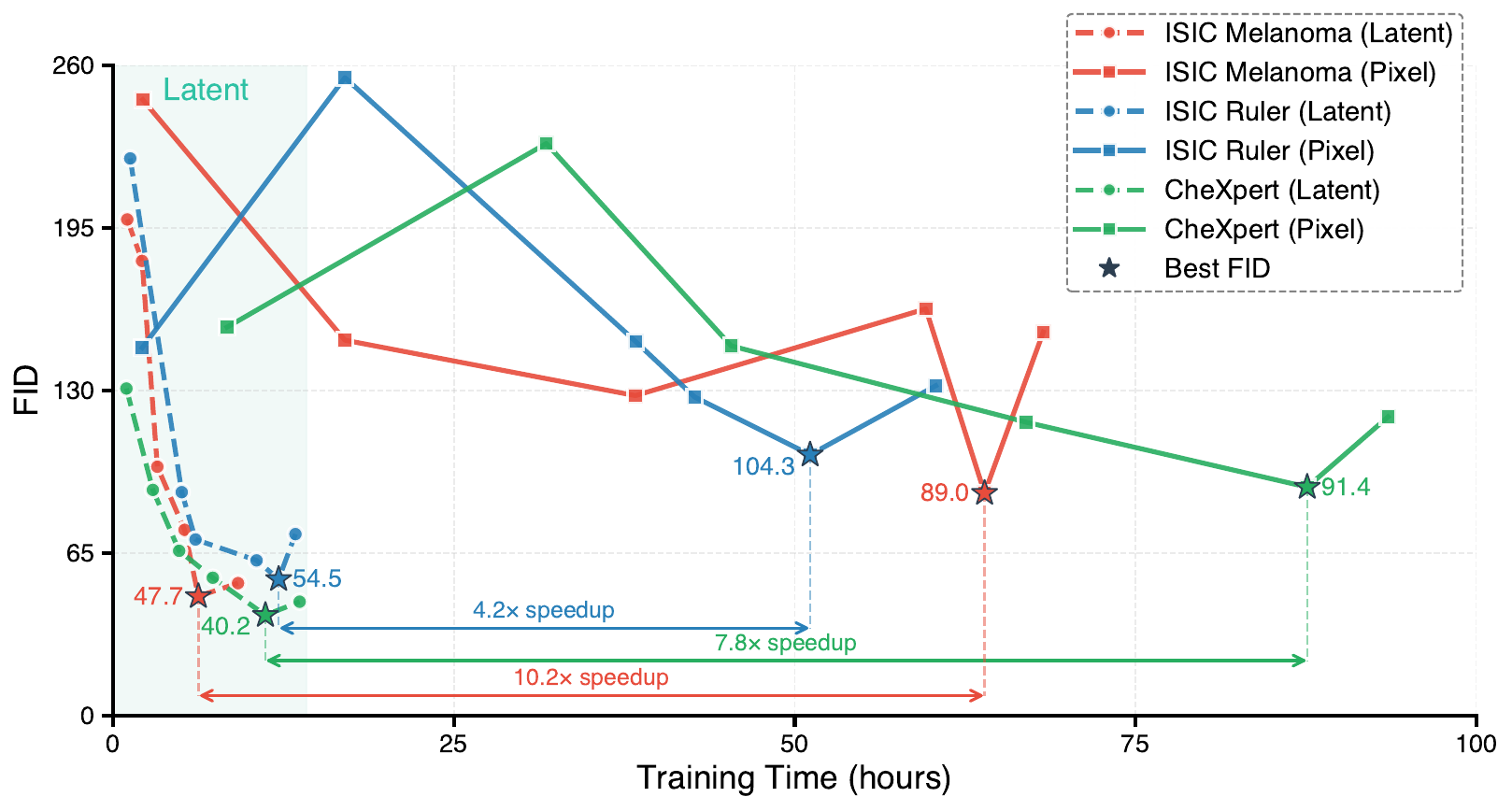}
    \end{subfigure}%

    \caption{
\textbf{Training efficiency comparison.}
\textbf{Left:} Classifier training in latent and pixel spaces, validated by AUROC over time. 
\textbf{Right:} Diffusion model training in latent and pixel spaces, validated by FID over time.
}
    \label{fig:trainingefficiency}
\end{figure}

\noindent \textbf{Pipeline-Level Efficiency Comparison.} To quantify efficiency gains, we evaluate the entire counterfactual pipeline, including classifier training, diffusion training, and counterfactual inference.

As shown in Fig.~\ref{fig:trainingefficiency} (left), the latent-space classifier converges 3.0$\times$--6.8$\times$ faster while maintaining comparable or improved validation AUROC.
During diffusion training (Fig.~\ref{fig:trainingefficiency} (right)), unconditional samples are periodically generated to compute FID against the validation set. The latent-space model converges 4.2$\times$--10.2$\times$ faster and achieves lower FID, eliminating the 40--80 hour training cost of pixel-space diffusion. More importantly, during counterfactual generation (Fig.~\ref{fig:pipeline_efficiency}), inference time is reduced by 2.5$\times$, 4$\times$, and over 40$\times$ compared to MaskDiME, FastDiME, and DiME, respectively, while peak GPU memory usage drops by 7$\times$, 13$\times$, and 10$\times$. MedDiME also attains the lowest FID, demonstrating that large-scale efficiency gains do not compromise generative quality.

\begin{table*}[t]
\centering
\fontsize{8}{9}\selectfont
\caption{
Quantitative results of counterfactual generation on ISIC2018.
Best and second-best results are shown in \textbf{bold} and \underline{underline}, respectively.
}
\label{tab:isic_results}
\setlength{\tabcolsep}{2pt}
\renewcommand{\arraystretch}{0.9}

\begin{tabular}{l|ccccc|ccccc}
\toprule

\multirow{2}{*}{Method} 
& \multicolumn{5}{c|}{\textbf{Melanoma--Nevus}} 
& \multicolumn{5}{c}{\textbf{Ruler--Non Ruler}} \\
\cmidrule(lr){2-6} \cmidrule(lr){7-11}

& FID\ensuremath{\downarrow} & L1\ensuremath{\downarrow} & MAD\ensuremath{\uparrow} & S$^3$\ensuremath{\uparrow} & FR\ensuremath{\uparrow}
& FID\ensuremath{\downarrow} & L1\ensuremath{\downarrow} & MAD\ensuremath{\uparrow} & S$^3$\ensuremath{\uparrow} & FR\ensuremath{\uparrow} \\
\midrule

DiME~\cite{jeanneret2022diffusion}
& 62.4 & 0.080 & \underline{0.611} & 0.911 & 91.8
& 55.5 & 0.086 & \textbf{0.864} & 0.944 & 90.1 \\

FastDiME \cite{weng2024fast}
& 36.9 & 0.051 & 0.594 & 0.947 & 95.3
& 41.1 & 0.073 & 0.837 & 0.963 & 89.9 \\

FastDiME-2 \cite{weng2024fast}
& 35.1 & 0.069 & 0.594 & 0.943 & 95.3
& 41.3 & 0.074 & 0.827 & 0.946 & \underline{91.3} \\

FastDiME-2+ \cite{weng2024fast}
& 38.8 & \underline{0.045} & 0.589 & \underline{0.964} & \underline{96.1}
& 47.2 & \textbf{0.061} & 0.810 & \underline{0.964} & 90.8 \\

MaskDiME \cite{guo2026maskdime}
& \underline{33.3} & \textbf{0.039} & 0.590 & 0.948 & 93.4
& \underline{34.2} & 0.084 & 0.808 & 0.922 & 91.0 \\

\textbf{MedDiME}
& \textbf{24.3} & 0.050 & \textbf{0.633} & \textbf{0.966} & \textbf{96.6}
& \textbf{28.5} & \underline{0.070} & \underline{0.843} & \textbf{0.965} & \textbf{91.8} \\

\bottomrule

\end{tabular}
\label{tab:isic2018}
\end{table*}

\begin{table}[t]
\centering

\begin{minipage}[t]{0.48\linewidth}
\centering
\fontsize{8}{9}\selectfont
\caption{
Quantitative results comparison on\textbf{ CheXpert}.
}
\label{tab:chexpert_results}
\setlength{\tabcolsep}{0.4pt}
\renewcommand{\arraystretch}{0.9}

\begin{tabular}{l|ccccc}
\toprule
Method
& FID\ensuremath{\downarrow} & L1\ensuremath{\downarrow} & MAD\ensuremath{\uparrow} & S$^3$\ensuremath{\uparrow} & FR\ensuremath{\uparrow} \\
\midrule
DiME
& 51.7 & 0.088 & 0.956 & 0.902 & 97.7 \\
FastDiME
& 29.2 & 0.058 & \underline{0.959} & 0.927 & \textbf{100.0} \\
FastDiME-2
& 30.4 & 0.069 & 0.953 & 0.881 & \underline{99.8} \\
FastDiME-2+
& 38.1 & \textbf{0.052} & 0.954 & 0.928 & \textbf{100.0} \\
MaskDiME
& \textbf{22.0} & \underline{0.053} & 0.954 & \underline{0.935} & \textbf{100.0} \\
\textbf{MedDiME}
& \underline{22.3} & \underline{0.053} & \textbf{0.961} & \textbf{0.940} & \textbf{100.0} \\
\bottomrule
\end{tabular}
\end{minipage}
\hfill
\begin{minipage}[t]{0.49\linewidth}
\centering
\fontsize{8}{9}\selectfont
\caption{
Ablation study of MedDiME on ISIC2018\textbf{ (Ruler--Non Ruler).}
}
\label{tab:ablation_topk}
\setlength{\tabcolsep}{0.55pt}
\renewcommand{\arraystretch}{0.97}

\begin{tabular}{lccccc}
\toprule
Method & FID\ensuremath{\downarrow} & L1\ensuremath{\downarrow} & MAD\ensuremath{\uparrow} & S$^3$\ensuremath{\uparrow} & FR\ensuremath{\uparrow} \\
\midrule
MaskDiME & 34.2 & 0.084 & 0.808 & \underline{0.922} & \underline{91.0} \\
\midrule
w/o Mask & 62.1 & 0.126 & 0.795 & 0.643 & 76.5 \\
Pixel Diff Mask & 44.7 & 0.090 & 0.805 & 0.797 & 88.9 \\
Fixed Mask & \textbf{26.3} & \underline{0.074} & \underline{0.812} & 0.883 & 90.2 \\
\textbf{MedDiME} & \underline{28.5} & \textbf{0.070} & \textbf{0.843} & \textbf{0.965} & \textbf{91.8}\\
\bottomrule
\end{tabular}
\end{minipage}

\end{table}

\noindent \textbf{Comparison with State-of-the-Art Methods} 
Quantitative results in Table~\ref{tab:isic2018} and Table~\ref{tab:chexpert_results} show that MedDiME achieves
the best or near-best performance across tasks. For realism, operating entirely in latent space mitigates high-frequency artifacts inherent to pixel-space diffusion~\cite{rombach2022high}, yielding the lowest or near-lowest FID. For proximity, classifier-gradient-driven masking restricts updates to decision-relevant regions while preserving the remaining latent structure~\cite{sobieski2025rethinking}. This leads to minimal or near-minimal pixel-level deviation ($L_1$) and the highest SimSiam similarity (S$^3$)~\cite{chen2021exploring}.
For validity, sharing the classifier gradient between diffusion guidance and spatial masking enables targeted movement toward the decision boundary without enlarging the editable region~\cite{dhariwal2021diffusion}. This results in the best Flip Rate (FR) and strong MAD~\cite{jeanneret2023ace}.

\begin{figure}[t]
  \centering
  \includegraphics[width=\textwidth]{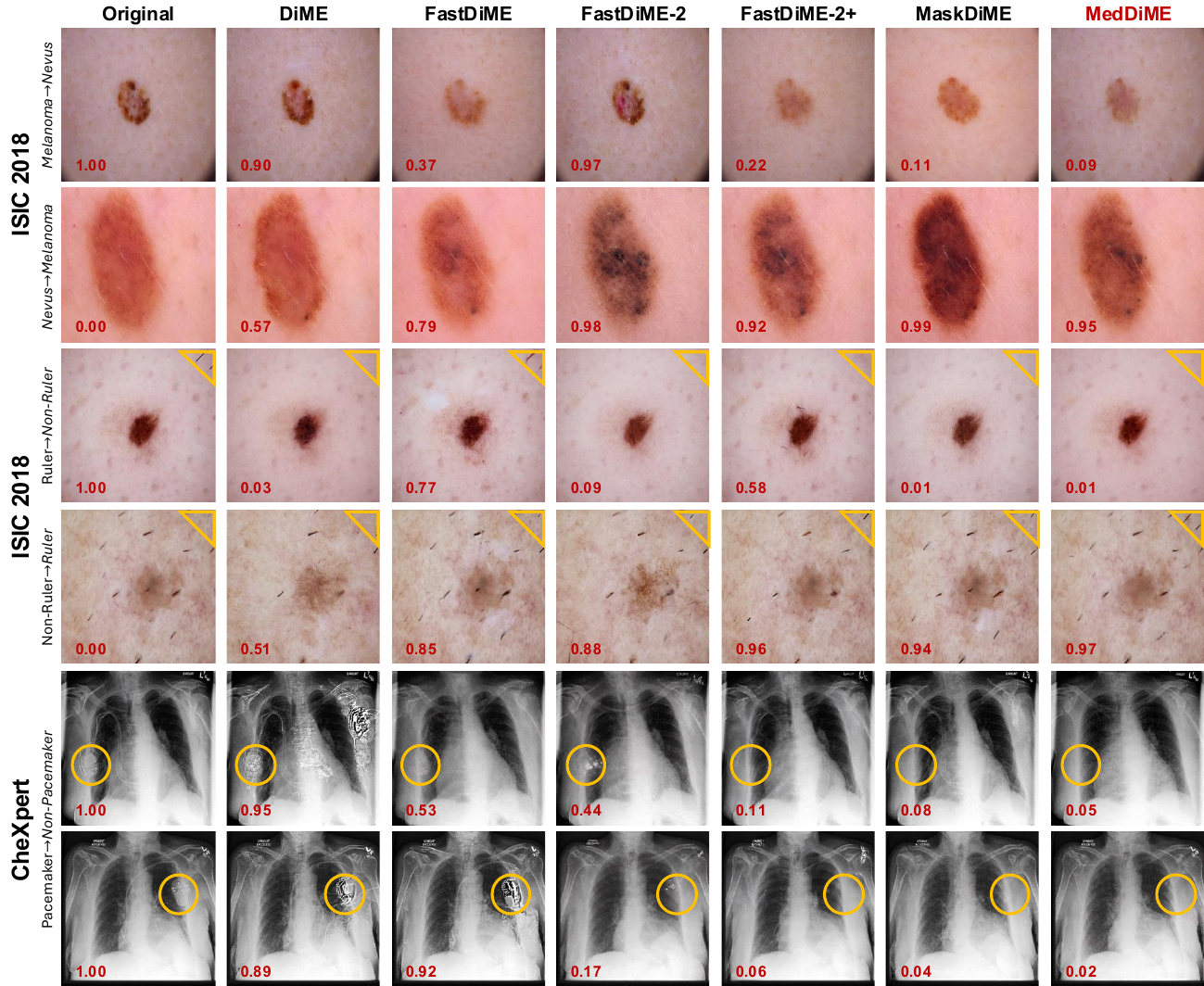}
\caption{\textbf{Qualitative counterfactual results across datasets.} 
Edited regions are highlighted in yellow. Numbers indicate predicted class probabilities 
(1 = Melanoma / Ruler / Pacemaker, 
0 = Nevus / Non-Ruler / Non-Pacemaker).}
  \label{fig:qualitative_results}
\end{figure}

Qualitative results in Fig.~\ref{fig:qualitative_results} further corroborate the quantitative results. In Melanoma$\leftrightarrow$Nevus transformations, edits are primarily confined to lesion regions. For salient but non-pathological visual markers (e.g., Ruler and Pacemaker tasks), modifications are restricted to the corresponding display structures, while surrounding non-diagnostic regions remain largely unchanged. In contrast, baseline methods~\cite{jeanneret2022diffusion,weng2024fast} tend to produce more dispersed edits, accompanied by additional background changes outside the target regions.

\noindent \textbf{Ablation Study.}
We compare different spatial masking strategies on ISIC2018 (Ruler--Non Ruler) (Table~\ref{tab:ablation_topk}). 
Without masking, gradient updates are applied globally, inevitably perturbing non-decision regions and degrading all metrics. 
Pixel Difference Mask imposes spatial constraints based on pixel discrepancies rather than classifier gradients, partially reducing global perturbations but failing to consistently localize decision-relevant regions, resulting in limited performance gains.
Fixed Mask further strengthens spatial restriction, improving FID and $L_1$ due to reduced edit magnitude and slightly increasing MAD; however, its static design cannot adapt to evolving gradient directions across diffusion steps, limiting its ability to maintain decision-aligned updates. 
In contrast, MedDiME dynamically aligns the editable region with classifier gradients at each reverse step, ensuring that updates remain spatially focused and decision-aligned, thereby achieving the strongest overall performance.

\section{Discussion and Conclusion}
We present MedDiME, a fully latent-space framework for classifier-guided medical counterfactual generation. By incorporating adaptive gradient-driven spatial masking, it substantially reduces computational and memory overhead without compromising counterfactual quality. Nevertheless, the structurally intricate and detail-dense nature of pacemakers makes realistic synthesis challenging for current diffusion models. Consistent with FastDiME~\cite{weng2024fast}, we therefore restrict evaluation on CheXpert to pacemaker removal. Furthermore, although $512 \times 512$ resolution and the adopted VAE sufficiently support the current experiments, future work will explore higher resolutions (e.g., $1024 \times 1024$) and medical-specific latent representation models to better capture fine-grained structural details.

\end{document}